\documentclass[10pt,twocolumn,letterpaper]{article}

\usepackage{cvpr}
\usepackage{times}
\usepackage{epsfig}
\usepackage{graphicx}
\usepackage{amsmath}
\usepackage{amssymb}

\usepackage[utf8]{inputenc} % allow utf-8 input
\usepackage[T1]{fontenc}    % use 8-bit T1 fonts
\usepackage{url}            % simple URL typesetting
\usepackage{booktabs}       % professional-quality tables
\usepackage{amsfonts}       % blackboard math symbols
\usepackage{nicefrac}       % compact symbols for 1/2, etc.
\usepackage{microtype}      % microtypography
\usepackage{graphicx}
\usepackage{placeins}
\usepackage{color}
\usepackage{subcaption}
\usepackage{amsmath}

\usepackage[pagebackref=true,breaklinks=true,letterpaper=true,colorlinks,bookmarks=false]{hyperref}

\cvprfinalcopy % *** Uncomment this line for the final submission
\def\cvprPaperID{21} % *** Enter the CVPR Paper ID here

\ifcvprfinal\fi

\begin{document}

%%%%%%%%% TITLE
\title{P2L: Predicting Transfer Learning \\ for Images and Semantic Relations}

\author{Bishwaranjan Bhattacharjee \textsuperscript{1}, John R. Kender \textsuperscript{2}, Matthew Hill \textsuperscript{1}, \\ 
Parijat Dube \textsuperscript{1}, Siyu Huo \textsuperscript{1}, Michael R. Glass \textsuperscript{1}, Brian Belgodere \textsuperscript{1}, \\ 
Sharath Pankanti \textsuperscript{1}, Noel Codella \textsuperscript{1}, Patrick Watson \textsuperscript{3} \\
\textsuperscript{1} IBM T.J. Watson Research Center, \textsuperscript{2} Columbia University, \textsuperscript{3} Minerva Project
}

\maketitle
%\thispagestyle{empty}

%%%%%%%%% ABSTRACT
\begin{abstract}
%Transfer learning enhances learning across tasks, by leveraging previously learned representations--if they are properly chosen.
We describe an efficient method to accurately estimate the effectiveness of a previously trained deep learning model for use in a new learning task. 
We use this method, 
% which we call 
``Predict To Learn'' (P2L), 
% in two very different domains, 
% visual and linguistic, 
to predict the most likely ``source'' dataset to produce effective transfer for training on a  ``target'' dataset. 
We validate our approach extensively 
% by assembling a collection of candidate source models, then fine-tuning each source candidate model to perform each of a collection of target tasks, and finally measuring how well transfer has been enhanced. 
across 21 tasks, including image classification tasks and semantic relationship prediction tasks in the linguistic domain. 
% within two different domains, 
The P2L approach selects the best transfer learning model on 62\% of the tasks, 
%while the alternative heuristic of choosing model trained with the largest data set selected the best model 
compared with a baseline of 48\% of cases when using a heuristic of selecting the largest source dataset and 52\% of cases when using a distance measure between source and target datasets.
%, effectively improving the state of the art of the efficacy of the source model prediction significantly (29\%)!
Further, our work
results in an 8\% reduction in error rate.
%also quantifies the impact of better model selection on performance: our method resulted in an overall 2.8\% (7.9\%) increase (decrease) in accuracy (error). These results suggest that P2L captures important common information  between source and target tasks, and that this is more effective in successful transfer learning than the conventional method of selecting source model based exclusively on dataset size alone. 
Finally, we also show that a model trained from merging multiple source model datasets  does not necessarily result in improved transfer learning.
% compared to those of individual source model datasets.  
This suggests that performance of the target model depends upon
% not merely on the presence similar/dissimilar data instances in the source dataset but also on 
the relative composition of the source dataset as well as their absolute scale, as measured by our novel method we term `P2L'.
\end{abstract}

\section{Introduction}

 In deep learning, large number of examples often help capture a robust representation of the unknown input distribution~\cite{Kavzoglu2012} since small data sets may not sufficiently sample the input space. However, in practice, small training jobs are common and labeled data is scarce in many domains. In a survey of industry visual recognition tasks, the users submitted on average 250 images comprising 5 labels per task  (see Section~\ref{sec:mergedDataset}). 

%To be clear, 
Our goal is not cross-task transfer. Our aim is to devise a practical and effective guideline for domain adaptation, for intra-task (such as image classification, or relationship prediction) cross-domain transfer, such as transfer from a classification model trained on a subset of ImageNet to a classification model for some unknown image classes related to a problem like industrial defect detection. A motivation is to optimize training of models in a cloud-based vision API.

\textit{Inductive transfer} learning methods \cite{QiangYang2010,Weiss2016} have been identified as a possible solution to this problem. These methods use knowledge acquired in a ``source'' task to enhance the learning of a new ``target'' task. However, these methods commonly assume that there is a ``best'' transfer model, typically the model trained with the largest data set \cite{DBLP:journals/corr/RazavianASC14}. Yet this assumption stands in tension with results showing that while a well chosen source can improve performance significantly, a poorly chosen source results in  worse performance than random initialization \cite{QiangYang2010,rosenstein2005transfer}. An open challenge remains: for fine-tuning of neural nets, 
on how to predict the effectiveness of transfer %between different source and target domains
prior to training. 

% MH: suggests these results belong in experiment section; also we have not defined what Tools or Sports subsets are yet.
%For example, when using ImageNet's "Tools" data classification as a target task, after the same amount of training, random initialization of provides 0.12 Top-1 accuracy. Pre-training on "Sports" reduces accuracy to 0.11, while pre-training on "Animals" increases final accuracy to 0.2. 
%maybe mention that these accuracies are based on a small training size
In this work, we describe a method for identifying good transfer models {\it prior} to training, that we then validate for commonly used ML tasks in both visual and linguistic domains. 
% This is valuable since 
A cloud based API which trains deep models for users, for example, must be prepared to train accurate models from widely varied target tasks automatically, while minimizing training time (and computational resources) and maximizing accuracy.
Precluding exhaustive search
%entailing fine-tuning all existing source models. 
at target task training time, P2L requires only a single forward pass of the target data set through a single reference model to identify, via a predictive algorithm, the most likely candidate for fine-tuning.
%Further, this pass only involves the relatively inexpensive effort of feature extraction, and does not rely on more expensive tasks of classification.
%MH I don't think classification is notably more expensive than feature extraction

In brief, beginning with a single reference model (for images, VGG16 trained on ImageNet1K, and for semantic relations PCNN), we first generate feature vectors for each source dataset. We then use these models to characterize the similarity between source domain features and the target domain's features.
Combining this similarity measure with a non-linear measure of source domain size results in a measure  that
reflects the source most likely to provide a useful embedding, independent of the larger reference model. 
% I DO NOT UNDERSTAND THE FOLLOWING SENTENCE! THIS LOOKS OUT OF PLACE HERE!  HELP! jrk
% Using this metric, we estimate the similarity between a conceptual category of inputs, and each member of our family of classifiers. We then fine-tune a network for each combination of source and target to assess the degree to which each of the source models enhanced learning. 

\section{Related Work}
\label{csofa}
Transfer learning literature explores a vast number of diverse strategies such as ensemble learning, co-training, model selection, collaborative filtering, few-shot learning \cite{bollegala2011relation} \cite{Socher2013}, domain adaptation \cite{Patricia2014}, weight synthesis \cite{Sussillo2012}, and multi-task learning \cite{jiang2009multi} \cite{nguyen2016} \cite{torrey2010transfer} and combinations thereof. Researchers have also investigated  practical considerations for domain transfer with limited or incomplete annotations~\cite{transfer3} and often suggesting  novel learning architectures and optimization objectives effective for such scenarios.

\textbf{Representation transfer:}
Representation transfer (RT) learning approaches share a common intuition that
% \cite{Bengio2011} SP 
compact representations learned from a ``source'' task can be reused to improve performance on a related ``target'' task.  \textit{Instance-based} approaches attempt to identify appropriate data used in the source task to supplement target task training, \textit{feature-representation} approaches attempt to leverage source task weight matrices, and \textit{parameter-transfer} approaches involve re-using the architecture or hyper-parameters of the source network~\cite{Dai2007,QiangYang2010}. These approaches, often supplemented by related small-data techniques such as bootstrapping, can yield improvements in performance (e.g.,~\cite{Azizpour2016}).
% \cite{Mou2016} \cite{nguyen2016} 
% \cite{vgg16} 
%\cite{transfer4}. 

Meta-learning~\cite{Lemke2015} is another approach for representation transfer. While meta-learning typically deals with training a base model on a variety of different learning tasks, transfer learning is about learning from multiple related learning tasks~\cite{FinnAL17}. Efficiency of transfer learning depends on the right source data selection, whereas meta-learning models could suffer from 'negative transfer~\cite{QiangYang2010} of knowledge if source and target domains are unrelated.
% SP just a distraction for our focus
% Surprisingly, in image classification performance gains are commonly observed even in cases where initialization data appears visually and semantically different from the target dataset (such as ImageNet and Medical Imaging datasets). 

%SP distraction
%In contrast, for relation prediction, semantic dissimilarity between source and  target task typically prevents effective transfer learning \cite{Mou2016}; consequently, the semantic-relations transfer is more poorly explored. However, semantic relations can contain information that can support transfer, one approach used vectorized representations of semantic relations as an added source of information to support image-segmentation \cite{Myeong2013} \cite{Rohrbach2010}. 

One approach to RT transfer learning is to leverage existing deep nets trained on  other large dataset(s), for example VGG16~\cite {DBLP:journals/corr/RazavianASC14} 
% \cite{vgg16} 
for images classification or PCNN~\cite{zeng2015distant} for relation prediction. The trained weights in these networks have captured a representation of the input that can be transferred by fine-tuning the weights or retraining the final dense layer of the network on the new task.  
Most DNN-based RT works assume there is only one source model, usually trained from ImageNet, whereas
% \cite{ImageNet22K}.  
P2L considers the problem of transfer learning when multiple source models are available. 

The Learning to Transfer~\cite{ying2018transfer} framework learns a \textit{reflection function} that transforms feature vector representations to be more effectively classified using a kNN-based approach. Although it uses a model trained on ImageNet to produce the initial feature vectors, it is not a parameter-transfer method, since the selected model is not fine-tuned on the target domains. Our experience is that it is difficult to curate a large number (tens) of prior experiences to adopt this approach in practice.
% and our focus in this study relies of availability of a few (tens) source models.  
% distraction SP
%% Additionally, such approach involves a number of meta-learning decisions, although in general each change from the original source architecture tends to decrease resulting classification performance \cite{transfer4}. %MG(reworked to below) Surprisingly, even in cases where initialization data appears visually and semantically different from the target dataset (such as ImageNet and Medical Imaging datasets), performance gains are commonly observed in comparison to training from random weights.

%MH: could use a cite of a medical imaging results up here ^^^^ 
\textbf{Fine-tuning variations:}
Our approach is most similar to that of selective joint fine-tuning~\cite{Ge}. The selective fine-tuning methods typically begin by using low-level features to identify images within a source dataset having similar low-level "textures" to a target dataset. Selective joint fine-tuning concludes by using a multi-task objective to fine-tune the target task using these images. A related approach has been used to enhance performance and reduce training time in document classification~\cite{Das2018} and to identify  examples to supplement training data~\cite{Ge,transfer4}. Our goal is to extend this approach to high-level features, and to domains outside computer vision, in order to construct a more complete map of the feature space of a trained network. In this aspect our work has some parallels with ``learning to transfer'' approaches~\cite{DBLP:journals/corr/abs-1708-05629}, but it attempts to train a source model optimized for transfer, rather than for target accuracy. 

While some recent studies in limited domains have related efficacy of this approach to a similarity between target and source datasets~\cite{transfer4}, and to the diversity~\cite{DBLP:journals/corr/RuderP17} of the examples,
%in limited domains, we have not come across literature that has 
few have explored the nature of performance improvement across multiple modalities and across multiple domains in realistic and real world settings.
%, which is primary focus of our work.
Another recent work~\cite{dube2019automatic} uses similarity as a metric for selecting a combination of source models which can be subsequently used for automatically labelling wild data samples, in order to fine-tune source models for a target task.
% Apart from having a different emphasis,  from the limited results presented it is not clear how well the technique 
However, it is unclear how these methods perform beyond the few datasets used in their empirical work.

\textbf{Other Approaches:}
When transferring information captured by previous task-learning for a new task, it is important to take into account the nature of both tasks.  One promising approach involves use of recommender systems~(e.g., task2vec \cite{achille2019task2vec}) which identify models with similar latent-space representations of labeled data. 
% In an object-detection context \cite{wang2015model}, this approach has been used to select likely candidates for inclusion in an ensemble model for object recognition.
In multi-task visual learning, a model learned to estimate the similarity space of various visual tasks is used to estimate the degree to which models trained to perform these tasks might contribute to transfer
%on a novel task
\cite{DBLP:journals/corr/abs-1804-08328}. %The current paper 
Our work 
aims, in part, to combine the low compute cost of the former estimation technique with the enhanced performance of the latter transfer technique, by learning a novel method for selection among previously trained source models.  
%However, our goal is not cross-task transfer in a multi-task formulation.
% - for example, we are not trying to transfer a model learned for a depth estimation task to a classification task.
% Although within a task type, (such as image classification) we do sometimes refer to a source task and target task, our aim is to devise a heuristic for domain adaptation, for intra-task, cross-domain transfer, such as transfer from a classification model trained on ImageNet to a classification model for Oxford Flowers.  We show that P2L works for multiple domains and for 2 types of tasks: image classification and semantic relation prediction.  

\section{Methods}
\label{methods}

\subsection{Embedding Divergence}
% changed this title to introduce our term, and it also didn't feel very "theory"-ish
\label{methods:embed_div}

Our goal is to make an optimal choice among pre-trained network weights learned for a target task $t_i \in T= (t_1, ..., t_N)$ from source tasks $s_j \in S = (s_1, ..., s_M)$. 
Given a target task and dataset $t_i$, a model $M(t_i, s_j)$ is generated by first training on the source task and dataset $s_j$, and then this information is transferred to $t_i$ through mechanisms such as fine-tuning.  
For each pair $(t_i, s_j)$, performance improvement by transfer in each scenario can be measured by:

\begin{equation}
\label{eq:improvement}
I(t_i, s_j) =  P(M(t_i, s_j)) - P(M(t_i, \phi)) 
\end{equation}

where $P(\cdot)$ is some defined performance evaluation (such as accuracy), $\phi$ represents the null source task and dataset (that is, the model $M(t_i, \phi)$ uses randomly initialized weights), and $I(t_i, s_j)$ is the measured performance improvement.
Determining the optimal $s_j$ for $t_i$ would then be achieved by optimizing $I(t_i, s_j)$.

However, since exhaustively training every possible model for $t_i$ is computationally expensive, we build a reliable estimator for $I(\cdot, \cdot)$, whose optimum could be used instead to quickly choose the optimal $s_j$.
Based on extensive experimentation, we propose this estimator, $E(\cdot, \cdot)$, which we term the ``embedding divergence'', as:

\begin{equation}
\label{eq:estimate}
E(t_i, s_j) = z(log(|s_j|)) + k \cdot z(D(t_i, s_j))
\end{equation}

where $|s_j|$ is the size of the source dataset $s_j$, $D(\cdot, \cdot)$ is a computed ``distance'' between the target and the source datasets, and $k$ is a learned parameter.
The standard z-scaling function, defined as $z(x) = (x - \mu)/\sigma$ is not strictly necessary, but it makes it easier to compare and to display intermediate results.

\begin{figure*}[ht]
   \centering
   \includegraphics[width=6in]{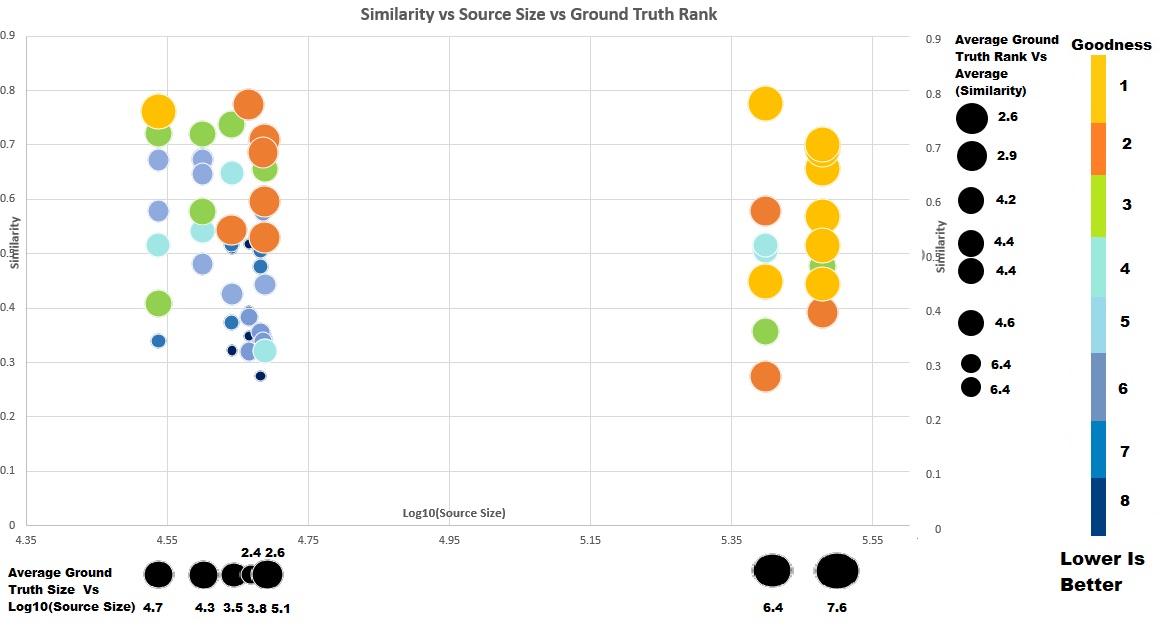}
   \caption{Relationship of performance of the target model to size of the source dataset ($X$-axis)  and its similarity with source dataset ($Y$-axis) for 9 targets over 8 sources. Both the similarity with the target and the size of the source dataset correlate with the performance of the target model. Color and size of each bubble reflect the performance of the target model.}
   \label{fig.heatdots}
\end{figure*}

The equation reflects the empirical observation that a larger source dataset tends to generate a more improved target model, and that this improvement tends to grow with size--but only logarithmically~\cite{Hestness2017}.  
 Importantly, the equation also reflects the empirical observation that size alone is an insufficient estimator, and that  dissimilarity between the datasets tends to be a negative factor that works against size. As shown in Figure~\ref{fig.heatdots}, for 9 target datasets over 8 sources (($S_v, T_v$) in section~\ref{gen_inst2}), we found that the performance of the target task is strongly correlated with both the similarity of the target dataset with the source, as well as the size of the source dataset itself.

%\begin{figure*}[ht]
 %  \centering
 %  \includegraphics[width=6in]{figures/goodness_gt3.jpg}
%   \caption{Relationship of performance of the target model to size of the source dataset ($X$-axis)  and its similarity with source dataset ($Y$-axis) for 9 targets over 8 sources. Both the similarity with the target and the size of the source dataset correlate with the performance of the target model. Color and size of each bubble reflect the performance of the target model.}
  % \label{fig.heatdots}
%\end{figure*}

We describe our choice of $D(\cdot, \cdot)$ based on two factors.  

First, we represent each dataset by a single, summarizing feature vector, $F(\cdot)$.  For example, in our experiments with labeled images,  $F(t_i)$ is computed from a convolutional neural network, by extracting for each image the vector produced at a specific layer of the network, and then summarizing the entire dataset by a statistical technique, such as a mean or a trimmed mean. Just as easily, $F(t_i)$ could be computed by another method, such as a future type of reference network, as long as it can represent the entire dataset.  

Second, the chosen dissimilarity function $D(t_i, s_j)$ must be shown to meaningfully compare the two vectors, being small for ``near'' datasets, and yet be meaningful for high-dimensional vectors. 
Candidates for $D(\cdot, \cdot)$ would include Kullback-Leibler or Jensen-Shannon divergence,
% (\cite {js_distance},
% \cite{kullback1951}), 
or Chi-square distance, or a Minkowski metric with $p=1$ (\textit{cityblock} distance) or  $p=2$ (\textit{Euclidean} distance). 
 
Once we have selected a choice of $D(\cdot, \cdot)$,  the value of $k$ can be tuned based on the performance of the approximation function $E(\cdot, \cdot)$ in comparison to the ground truth improvement function $I(\cdot)$. The value of $k$ is learned by first training $E(\cdot, \cdot)$ on a collection of target and source datasets, and then evaluating the quality of the estimation on a held-out set of target and source datasets. 

Because the exact numeric values of the estimate $E(\cdot, \cdot)$ are not directly comparable with the measures of improvement in $I(\cdot)$, ``quality'' is defined as how similarly $E(\cdot, \cdot)$ ranks the order of the datasets $s_j$, compared to the ground truth ranking given by $I(\cdot)$.  
There are a number of ways to define rank order similarity, many of them based on the non-parametric correlation method of Spearman $\rho$.  
We note, however, that it is not necessary to evaluate how well $E(\cdot, \cdot)$ orders the {\em entire} collection $S$ of possible source datasets $s_j$; 
usually we are only interested in how similarly $E(\cdot, \cdot)$ orders some topmost $T$ datasets.
% Therefore, $T$ can even be equal to $1$ (i.e., ``Top-1'' similarity).

\begin{figure}
  \centering
   \includegraphics[width=3in]{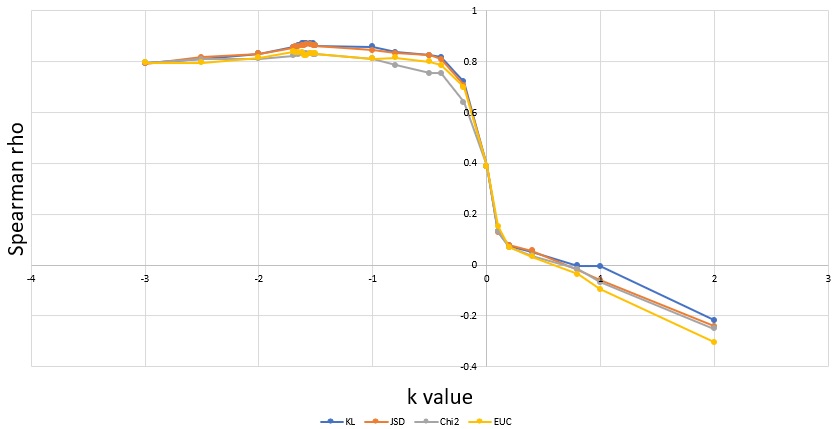}
  \caption{Relationship of correlation metric (Spearman $\rho$) with the balancing parameter $k$ and various similarity metrics (KL=KL Divergence, JSD=Jensen Shannon Distance, CHI2=$\chi^2$ distance, and EUC=Euclidean distance) studied.}
   \label{fig:ktune}
\end{figure}

In practice, we have found empirically that
(1) the second-last layer of a deep learning network gives good representative vectors for images, (2) averaging these vectors gives a good summary for a dataset, (3) that the choice of distance function is not critical--although KL-divergence and cityblock work well, and (4) that the results are not overly sensitive to the exact choice of $k$  (see Figure~\ref{fig:ktune}). 
The optimal choice of $T$ for measuring the quality of top-T ranking may depend on the statistical properties of the collection $S$, but it appears to be best if $T$ is small.

This work takes an engineering approach to proposing an approximation function $E(t_i, s_j)$.  
However, this framework is extensible to future work, which may explicitly learn other compact representations of the datasets, other inexpensive dissimilarity functions, and more sophisticated non-linear ways of modeling the observed interaction of size and distance.  

\subsection {Implementation Details for Images}
\label{impl_image}

\begin{figure}
   \centering
   \includegraphics[width=3in]{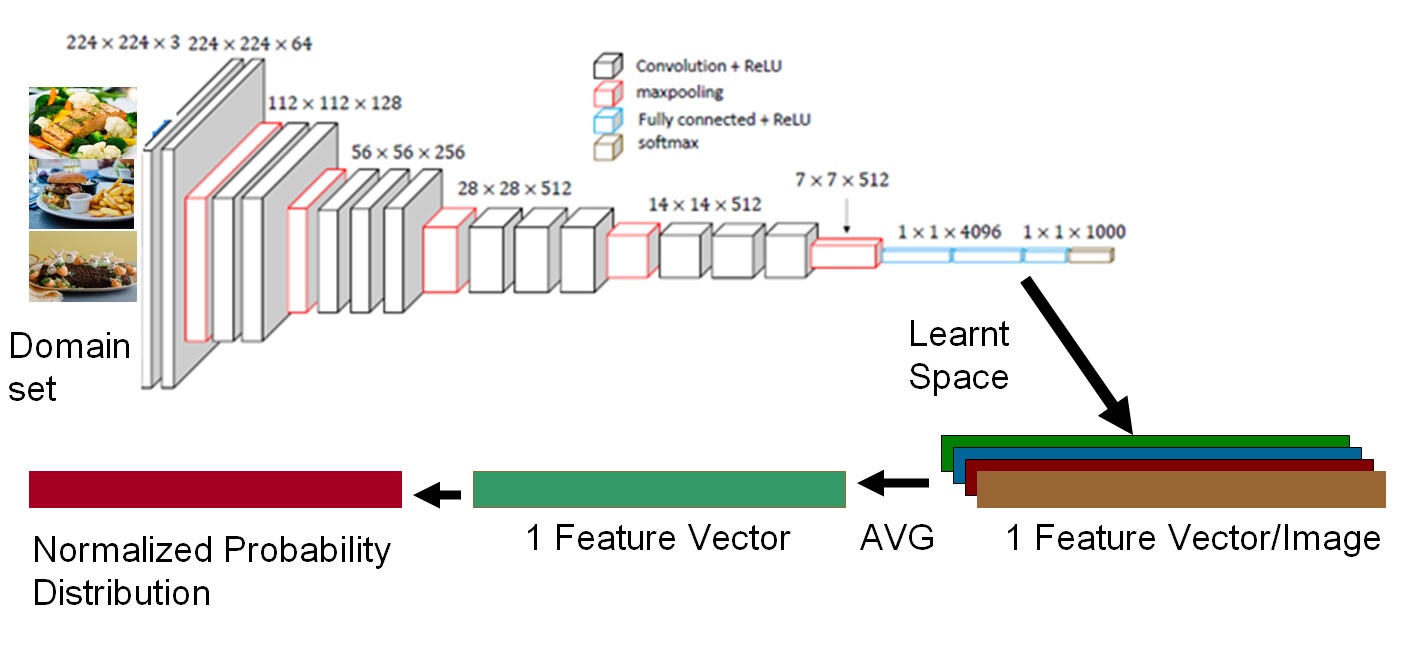}
   \caption{Deep Learning Pipeline for Images.}
   \label{fig:vision_pipeline}
\end{figure}

As described in Figure~\ref{fig:vision_pipeline}, we use as a reference model the VGG16 model pre-trained on ImageNet1K. 
For $F(\cdot)$, we first extract the response of the penultimate full connection layer, a 4096-dimensional vector. 
In a learning task with $k$ images, we extract $k$ such vectors $v_i$, compute their mean, $v_\mu$, and then L1-normalize this mean, giving $\overline{v_\mu}$ as the summary feature vector for this task. 
For $D(\cdot, \cdot)$, we compute one of several possible distance measures, smoothing any zero components by adding an appropriate $\epsilon$ value.

\subsection{Implementation Details for Semantic Relations}
\label{gen_inst1}
\newcommand{\kbent}[1]{\textsc{#1}}
\newcommand{\kbrel}[1]{\textit{#1}}
\newcommand{\kbtrip}[3]{$\langle \kbent{#1}\ \kbrel{#2}\ \kbent{#3} \rangle$}

\begin{figure}
	\includegraphics[width=0.83\linewidth]{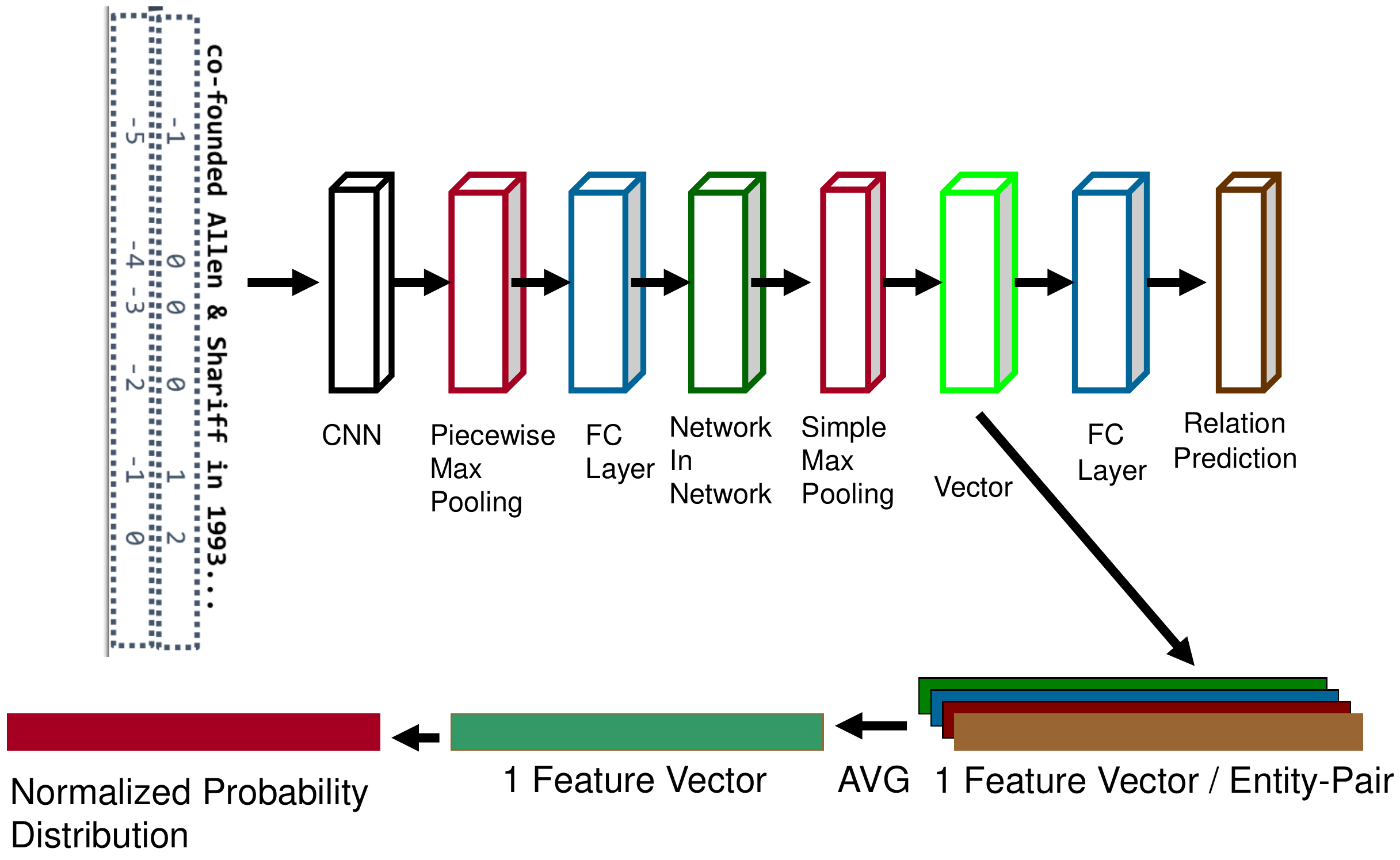}
   \caption{Deep Learning Pipeline for Relation Prediction.}
   \label{fig.rkiarch}
%\captionof{table}{Hyperparameters used}\label{tbl.rkihypers}
%\centering
%\begin{tabular}{rr}
%\toprule 
%\bf Hyperparameter & \bf Value \\ \midrule
%word dim. & 50 \\
%position dim. & 5 \\
%sentence vector & 400 \\
%NiN filters & 400 \\
%CNN filters & 1000 \\
%CNN filter width & 3 \\
%dropout & 0.5  \\
%learnRate & 0.003 \\
%decay multiplier & 0.99 \\ %This is only for training the target models
%batch size & 16 \\
%optimizer & SGD \\
%minLearnRate = 1.0E-4
%\bottomrule
%\end{tabular}

\end{figure}

The task of relation prediction provides a second benchmark for source domain selection. 
In this task, a semantic relations base is extended with information extracted from text. 
We use the CC-DBP~\cite{ccdbp} dataset: the text of Common Crawl\footnote{\url{http://commoncrawl.org}} and the semantic relations schema and training data from DBpedia~\cite{dbpedia}. DBpedia is a knowledge graph extracted from the infoboxes from Wikipedia. An example edge in the DBpedia knowledge graph is \kbtrip{Larry McCray}{genre}{Blues}, meaning Larry McCray is a blues musician.
This relationship is expressed through the DBpedia \kbrel{genre} relation, a sub-relation of the high level relation \kbrel{isClassifiedBy}. 
The relation prediction task is to predict the relations, if any, between two nodes in the knowledge graph from the entire set of textual evidence, rather than from each sentence separately as in mention-level relation extraction. %For example, \kbent{Caribou Coffee} and \kbent{Minnesota} are connected by the \kbrel{location} relation, a fact strongly indicated by the contexts in which they co-occur: 
Figure \ref{fig.rkiarch} shows the relation prediction neural architecture. %The architecture is typical of natural language processing and knowledge induction.
The feature representations are taken from the penultimate layer, the max-pooled network-in-network. All models have the same architecture and hyperparameters.
%,shown in Table \ref{tbl.rkihypers}
%relation prediction neural architecture

%\begin{tabular}{rrrrr}
%\toprule
%\bf Hyperparameter  & Value &    & Hyperparameter & Value \\
%\midrule
%word dim. &	50 &  &   NiN filters &	400 \\	       
%position dim. &	5  &  &  CNN filters &	1000 \\       
%sentence vector & 400 & &	    CNN filter width &	3 \\	       
%dropout & 0.5	  \\
%\bottomrule
%\end{tabular}

\section{Experimental Results and Analysis}
\label{gen_inst2}
\subsection{Experimental Approach: Images}

For evaluating P2L we used Caltech-UCSD Birds (CUBS)~\cite{cubs2011}, Stanford Cars (Cars)~\cite{cars2013}, Sketches~\cite{sketches2012}, Wikiart~\cite{wikiart2015} and Oxford Flowers (Oxford)~\cite{flowers2008} and 9 datasets from the Visual Decathlon Challenge~\cite{deca2017}: Aircraft, CIFAR100, Daimler Pedestrian Classification (Daimler), Describable Textures (DTD), German Traffic Sign (GTSRB), Omniglot, Street View House Number (SVHN), UCF101, VGG-Flowers.  These 14 datasets, representing fine grained classification tasks, serve as  targets ($T_v$) for our evaluation.  

\begin{figure}
   \centering
   \includegraphics[width=3in]{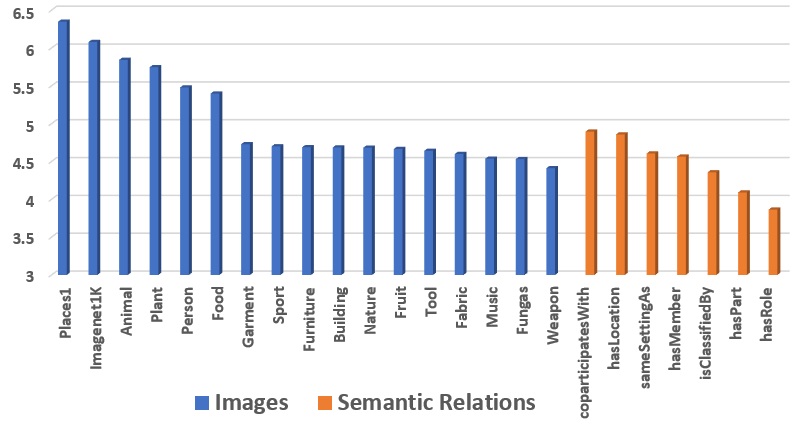}
   \caption{$Log_{10}(Size)$ of sources $T_v$ }
   \label{fig:size}
\end{figure}

For source datasets ($S_v$), we used  Places1~\cite{zhouplaces1}, ImageNet1K~\cite{IM1K} and 15 subsets from ImageNet22K~\cite{IM22K}. ImageNet22K contains 21,841 categories spread across hierarchical categories such as person, animal, fungus etc. We extracted some of the major hierarchies  from ImageNet22K (Fig \ref{fig:size}) to form multiple source image sets for our evaluation. A total of $\sim$9 million images were used. Some of the domains like animal, plant, person, and food contained substantially more images (and labels) than others categories such as weapon, tool, or sport . This skew is reflective of real world situations, and provides a natural test bed
%for our method 
when comparing training sets of different sizes. This is visible in Fig \ref{fig:size}

Each of these ImageNet22K domains was then split into four equal partitions. The first was used to train the source model, and the second was used to validate the source model. One-tenth of the third partition was used to create a transfer learning target and the fourth partition was used to validate it.  For example, the person hierarchy has more than one million images. This was split into four equal partitions of more than 250K each. The source model was trained with data of that size, whereas the target model was fine-tuned with one-tenth of that data size taken from one of the partitions. These smaller target datasets are reflective of real transfer learning tasks. We thus generated 15 source training datasets and 15 possible target training datasets from ImageNet22K. The 15 source datasets were used, along with Places1 and ImageNet1K, as source datasets for transfer learning.  

To tune $k$ in our approximation function $E(\cdot, \cdot)$ ( Equation~\ref{eq:estimate}), as well as to determine which dissimilarity measure to use, 9 of the target training tasks generated from ImageNet22K were used as a  training group, ($S_t, T_t$). These consisted of furniture, food, person, nature, music, fruit, fabric, tool, and building. The $k$ value thus generated was used for evaluation on the workloads in the 14 target tasks (referred to above as $T_v$), as well as 7 tasks for Semantic Relations. 

The training of the source and target models was done using Caffe  using a ResNet-27 model. 
% \cite {resnet27}. SP 
The source models were trained using SGD as in \cite {SGD} for 900,000 iterations with a step size of 300,000 iterations and an initial learning rate of 0.01. 
The target models were trained with an identical network architecture, but with a training method with one-tenth of both iterations and step size. A fixed random seed was used throughout all training. 

\subsection{Experimental Approach: Semantic Relations}
We split the task of relation prediction into seven subtasks composed of the high-level relations with the most positive examples in the CC-DBP; other relations were discarded. This was intended to mirror the partitions of ImageNet by high-level classes. The seven source domains ($S_v$) are shown in Figure~\ref{fig:size}.
A model is trained for each of these domains on the full training data for the relevant relation types.

Our approach to transfer learning was the same as in images: a deep neural network trained on the source domain was fine-tuned on the target domain. Fine-tuning involves re-initializing and re-sizing the final layer, since different domains have different numbers of relations. The final layer is updated at the full learn rate $\alpha$, while the previous layers are updated at $f\cdot\alpha$, with $f<1$. We used a fine-tune multiplier of $f=0.1$.

%selection of small training
A new, small training set is built for each target task. 
For each split of CC-DBP, we take 20 positive examples for each relation from the full training set. (If the total examples is fewer than 20, we take all the training examples.) We then sample ten times as many negatives (i.e., unrelated pairs of entities). These form the target training sets ($T_v$). The model trained from the full training data of each of the different subtasks is then fine-tuned on the target domain. We measure the area under the precision-recall (PR) curve for each trained model. We also measure the area under the PR curve for a model trained without transfer learning. %The performance of the transfer learned model divided by the performance of the model without transfer learning is the improvement.

\subsection{Results}
When training a model, we compare our P2L method against five baseline methods of initializing training weights:\\
	1. Source model trained on largest dataset (B1) like in \cite{Hestness2017}\\
	2. Source model trained on ImageNet1K (B2) for images.\\
	3. Randomly chosen source model from set of models (B3).\\
	4. No transfer learning: weights initialized randomly (B4).\\
	5. Source model trained on least divergent dataset (B5) \\
%\begin{enumerate}
%	\item choose the source model trained with the largest amount of data (B1);
%	\item use weights learned from ImageNet1K (B2) for images;
%	\item randomly choose a model from the set of available models (B3); and
%	\item not use transfer learning at all but instead initialize the weights of the network randomly (B4).
%\end{enumerate}
 We have used this to compare P2L across two domains: Images (Section~\ref{images}) and Semantic Relations (Section~\ref{dbpedia}).

In summary, as shown in Table \ref{fig:resultsum} across 21 tasks in the above two contexts,  P2L was able to deliver an average accuracy of 67.22\% compared to 64.47\% and 64.86\%for the baseline method of picking the largest dataset (B1) and most convergent (B5). Additionally, P2L was able to pick a better model in 13 out of 21 tasks. In three tasks where it did not pick the best, the prediction scores between its pick and the winner were very close. The Spearman correlation between the ground truth and the predictions over all the possible source datasets was strong for images (0.707) as shown in Table~\ref{fig:Spearman_for_measures} and semantic relations (0.763)
as shown in Fig~\ref{fig:combined2}.

Tables~\ref{fig:gain} and \ref{fig:SRgain} show the relative increase in final performance for our proposed method in comparison to each of these three methods, across images and relations. 
%(in Section~\ref{images}) and DBpedia (in Section~\ref{dbpedia}).  
In the case of images, we present a comparison against ImageNet1K also in Table~\ref{fig:gain}, since ImageNet1K is often used as a source dataset for transfer learning.

\begin{figure}[ht]
  \centering
   \includegraphics[width=3.5in]{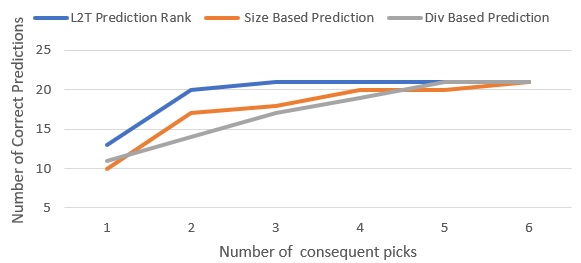}
   \caption{Number of attempts needed to get to best result for all 21 datasets across image and semantic relations}
  \label{fig.num_picks}
\end{figure}

Additionally in Figure \ref{fig.num_picks}, we see that P2L is able to pick the best source model for the 21 tasks in maximum of 3 picks from the basket of source models. In contrast, the largest dataset method (B1) would take 6 picks and the most convergent (B5) would take 5 picks.

The latency of a prediction is low. For a dataset like DTD it was 65 seconds on a K40 GPU end to end. In contrast, a training run on DTD takes ~43200 secs on the GPU. The prediction latency is largely a function of the number of images in the target dataset. Processing speed for a prediction over 17 source datasets was 30 images/second.

\subsubsection{Validation on Image space}
\label{images}
%ImageNet22K

%{\bf NOEL 5-17-2018: Guys, $T_{tr}$, $T_{te}$, $S_{tr}$, and $S_{te}$ must be defined in this section. Number of categories, number of Images, splits (25\%), etc ...}  

%\begin{figure*}[htb]
%   \centering
%	\begin{minipage}{.5\textwidth}
%	\centering
%   \includegraphics[width=.95\linewidth]{figures/ImageNet22K1jpg.jpg}
%   \caption{ImageNet22K partitions}
%  \label{fig.ImageNet22K}
%	\end{minipage}%
%\begin{minipage}{.5\textwidth}
%  \centering
%   \includegraphics[width=.9\linewidth]{figures/average_spearman_rha.png}
%   \caption{Spearman rank correlation for various measures}
%   \label{fig.Spearman_average}
%	\end{minipage}
%\end{figure*}

We tested distance measures based on Kullback-Leibler Divergence (KLD), Jensen-Shannon Divergence (JSD), Chi-square (CHI2), and Euclidean distance (ED). 
For each training task in ($S_t,T_t$), we calculated the rank-correlation (Spearman $\rho$) between the predictions of each of these measures, and the ground-truth transfer performance based on Top-1 classification accuracy. This is shown in Figure\ref{fig:ktune} 

% Figure \ref{fig.Spearman_average} shows the average Spearman $\rho$ of the Top-1 ground truth rank and our prediction rank as they varied with various $\alpha$ values of equation~(\ref{eq:predictionFunc}).
% For $\alpha$ in this interval, the KLD based measure is most sensitive, and we use it exclusively for evaluations with $\alpha_d$ = 1 and $\alpha_z$ = 4. 
% The parameter $\alpha_d$ was fixed at 1, since what is important is the ratio of $\alpha_z$ to $\alpha_d$. For these parameters, the average Spearman $\rho$ for the transfer learning task is 0.83. 
% The gains from our prediction method are shown in Table \ref{fig.gain}.

% The parameters $\alpha_d$ and $\alpha_z$  which were formulated as described above on the 6x6 ImageNet22K training set ($S_t, T_t$), were subsequently used for validation on the 9x9 ImageNet22K (in 4.3.1), real world image classification tasks (in 4.3.4), Oxford Flowers (in 4.3.3), as well as dbpedia (in 4.3.2) datasets.  
% The size of ($S_t, T_t$) can be increased from the current 6x6.  
% But even with the current size it shows the potential of P2L.

This parameter selection of $k$ % $\alpha_d$ and $\alpha_z$ 
is essentially offline, and only needs to be done once.
%to pick the parameter values. 
The same value of $k$ is used for images as well as semantic relations.
%in Section~\ref{dbpedia}.
% It required 30 custom training jobs.  
% All subsequent predictions for the 9x9 ImageNet22K, real work image classification tasks, Oxford Flowers, as well as dbpedia, did not require further training. 

\begin{table*}
\center
\caption{Summary of  Results}
\label{fig:resultsum}
\begin{tabular}{rrrrrrr}
\toprule
Domain & \multicolumn{3}{c}{Mean Top-1 Accuracy} &  \multicolumn{3}{c}{Mean Spearman Correlation}  \\
  & P2L   &   Largest & Least & P2L &  Largest & Least\\
  & (ours) & Dataset & Divergent & (ours) & Dataset & Divergent\\
\midrule
Images (14 tasks)	&	\bf 65.57 &  61.40 & 64.11 & \bf 0.703 & 0.685 & 0.532  \\  
Semantic Relations (7 tasks)	&	\bf 71.79 &  70.6 & 66.36 & \bf0.763 & 0.714  & 0.037	\\
\midrule
Average over 21 tasks & \bf 67.22   &  64.47 & 64.86 & \bf 0.723 & 0.695 & 0.367\\
\midrule
\end{tabular}
\end{table*}

\begin{table}
\center
\caption{Spearman $\rho$ for predictions vs ground truth for transfer learning on images using P2L}
\label{fig:Spearman_for_measures}
\begin{tabular}{rrrrr}
\toprule
Target & Spearman &    & Target & Spearman \\
Dataset & $\rho$ &    & Dataset & $\rho$\\
\midrule
CUBS &	0.821 &  &   Sketches &	0.843 \\	       
VGG-Flowers &	0.630  &  &  Daimler &	0.652 \\       
UCF101 & 0.777 & &	     Omniglot &	0.525 \\	       
Oxford & 0.718	 & &		     GTSRB &	0.520   \\
Aircraft &	0.608 &  &    SVHN  &   0.603   \\
DTD & 0.951  &&  Wikiart &  0.8407\\
Cars  &   0.64951   && CIFAR100 & 0.730
%\bottomrule
\end{tabular}
\end{table}

\subsubsection{Validation on Common Crawl - DBpedia}
\label{dbpedia}

\begin{figure}[ht]
  \centering
   \includegraphics[width=3.5in]{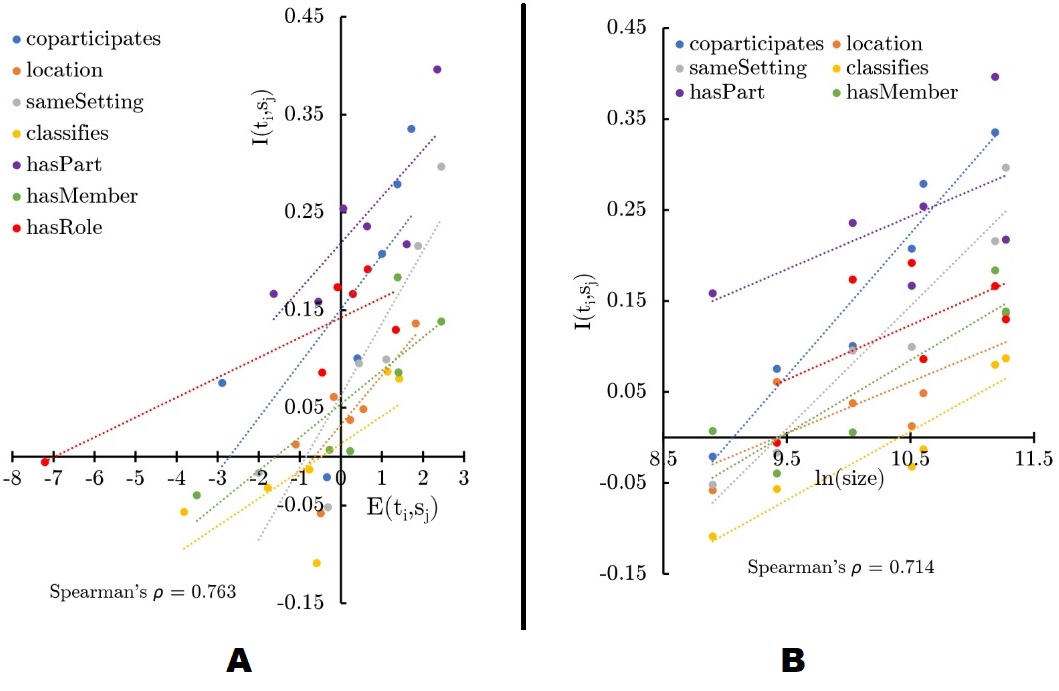}
   \caption{Transfer learning improvement for semantic relations. A: Predicted by KLD and size in CC-DBP. B: Predicted by size only in CC-DBP.}
  \label{fig:combined2}
\end{figure}

Figure \ref{fig:combined2}A shows the correlation of the prediction $E(t_i, s_j)$ with the improvement $I(t_i, s_j)$, when using KLD in addition to sizes of the source domains' training set in $E(\cdot, \cdot)$. Figure \ref{fig:combined2}B shows the same when only size is used. Using the estimator produced better predictions, that is, $E(t_i, s_j)$ and $I(t_i, s_j)$ were then better correlated (Spearman $\rho$ = .763).  Additionally the overall accuracy obtained using P2L at 71.79\%  was higher than the overall accuracy obtained using just size at 70.6\%.

\begin{table*}
\centering
%\captionsetup{justification=centering}
\caption{Gain Summary for Images. \textbf{P2L=} Accuracy using our method; \textbf{B1 =} Accuracy when largest source training dataset $S_t$ was used; \textbf{B2 =} Accuracy when training from ImageNet1K pretrained weights; \textbf{B3 =} Avg accuracy of randomly picked source dataset $S_t$; \textbf{B4 = } Accuracy achieved when training from random weights $M(t_i, \phi)$; \textbf{B5=} Least Divergent D($t_i$,$S_j$)
%IM = Accuracy using ImageNet1K has a source training dataset \\
%L2T = Prediction score of L2T pick (ours);     BP = Prediction score of best } %no transfer learning  $\phi$}
%$P_{P2L}$ =  Performance using P2L;  $P_{\hat{s}}$ = Performance using source with largest training dataset ;\\
%$\bar{P_{s}}$ = Average performance over all source datasets; $P_{\phi}$ = Performance of no transfer learning
}
\label{fig:gain}
\begin{tabular}{rrrrrrr}
\toprule
Target Dataset & P2L Picked   & Largest Training & Least & ImageNet1K & Random Dataset & No Transfer \\
 $t_i$ & Best  &  Dataset  & Divergent & Selection & Selection  & Learning \\
& Dataset ? & (P2L-B1)/B1 & (P2L-B5)/B5 &(P2L-B2)/B2 &  (P2L-B3)/B3 & (P2L-B4)/B4 \\
%& & Dataset ? &  $(P_{P2L}-P_{\hat{s}})/P_{\hat{s}}$ & $(P_{P2L}-\bar{P_{s}})/\bar{P_{s}}$ & $(P_{P2L}-P_{\phi})/P_{\phi}$  \\
\midrule
CUBS & \bf Yes  &1.00 & 0.00 &0.28 & 1.69 & 4.28 \\
VGG-Flowers & \bf Yes &   0.57 & 0.00 &0.21 & 0.82 & 2.13 \\
UCF101 &\bf	Yes &0.00 & 0.00 & 0.08 & 0.47 & 1.77  \\
Oxford & \bf Yes & 0.19 & 0.00 & 0.08  & 0.58  & 1.30\\
Aircraft & \bf Yes  &	0.00 & 0.00 & 0.01 &  0.67&  0.55\\
Sketches & \bf Yes &	0.08 & 0.17 & 0.00 & 0.17  &	7.26 \\
Daimler & \bf Yes  & 0.00 & 0.00 & 0.00 & 0.00 &   0.01 \\
Omniglot& \bf Yes  & 0.00 & 0.00 & 0.00 & 0.07  & 0.27  \\
GTSRB& \bf Yes & 0.00 & 0.00 & 0.00 & 0.00 & 0.01 \\
SVHN&	No  & -0.01 & 0.00 & 0.00& 0.02 & 0.02 \\
DTD & No  &0.00 & 0.60& -0.03  & 0.55 &	2.25 \\
Wikiart &No  & 0.00 & -0.04 &  -0.04 & 0.79 & 1.31 \\
Cars&  No  &	0.00 & 0.00 &-0.04  & 0.52&  2.56 \\
CIFAR100 &  No &	0.00 & -0.06 & -0.06 & 0.30  &	4.20 \\
\midrule
\end{tabular}
\end{table*}

\begin{table*}
\center
\caption{Gain Summary for Semantic Relations \\}
\label{fig:SRgain}
\begin{tabular}{rrrrrrr}
\toprule
Target Dataset & P2L Picked   & Largest Training  & Random Dataset & No Transfer \\
$t_i$ & Best Dataset?  &  Dataset  & Selection  & Learning \\
& &  (L2T-B1)/B1 & (P2L-B3)/B3 & (P2L-B4)/B4 \\
%& & Dataset ? &  $(P_{P2L}-P_{\hat{s}})/P_{\hat{s}}$ & $(P_{P2L}-\bar{P_{s}})/\bar{P_{s}}$ & $(P_{P2L}-P_{\phi})/P_{\phi}$  \\
\midrule

% old formula and new formula k=-1.5
% Note hasRole and hasMember are the best of the methods considered in the table but not the absolute best. I think the intention is actually to mean best overall so I changed them to 'No'
hasPart & \bf Yes & 0.15&  0.13 &  0.40\\
copartWith & \bf Yes &  0.00 & 0.14 & 0.34\\
sameSettingAs & \bf Yes &  0.00&  0.17 & 0.30\\
hasLocation & \bf Yes &  0.00& 0.09 & 0.14\\
hasMember & No  &  0.00 &  0.07 & 0.14\\
hasRole & No  &  0.00 &  0.01 & 0.13\\
isClassifiedBy & No &  -0.01&  0.08 & 0.08\\
%\bottomrule
% new formula k=-0.15
%Semantic &  hasPart & No & 0.00 &  -0.02 &  0.22\\
%Relations & copartWith & \bf Yes &  0.00 & 0.14 & 0.34\\
%& sameSettingAs & \bf Yes &  0.00&  0.17 & 0.30\\
%& hasLocation & \bf Yes &  0.00& 0.09 & 0.14\\
%& hasMember & No &  0.00 &  0.07 & 0.14\\
%& hasRole &  No &  0.00 &  0.01 & 0.13\\
%& isClassifiedBy & \bf Yes &  0.00 &  0.10 & 0.09\\
\midrule
\end{tabular}
\end{table*}

%\begin{figure}[ht]
%   \centering
%   \includegraphics[width=3.5in]{figures/newsize_im_vs_c1.jpg}
    % \includegraphics[width=4in]{figures/newsize_im_c1.jpg}
%   \caption{Accuracy of ImageNet1K vs. Combined dataset).}
%   \label{fig:notjustsize}
%\end{figure}

%\begin{figure}[ht]
%   \centering
 %  \includegraphics[width=3.5in]{figures/combined2.jpg}
    % \includegraphics[width=4.7in]{figures/newsize_im_c1.jpg}
 %  \caption{Accuracy vs Learning Rate Scales}
 %  \label{fig:combined2}
%\end{figure}

\subsubsection{Comparing Against Merged Source Datasets}
\label{sec:mergedDataset}
To help put these results in context,
we have investigated how well a merged dataset of various source domains could do in comparison to its individual components. 
While it may seem that a single merged dataset would perform as well or better than individual sources, in reality we have noticed results to the contrary.

We built a combined dataset of  ImageNet22K and Places2 into one large dataset (referred to as LC) and trained a ResNet27 model with it.  We then took 71 training datasets submitted by users to a custom learning cloud API, and performed transfer learning experiments from models trained on LC and ImageNet1K.  Given that ImageNet1K is a subset of ImageNet22K,  it is a subset of LC too.  The transfer experiments were done using 8 different learning rate regimes.

\begin{figure}
   \centering
   \includegraphics[width=3.5in]{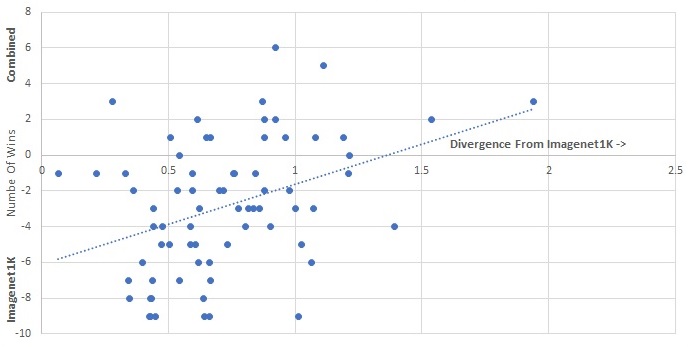}
   \caption{Accuracy: ImageNet1K vs. Combined dataset}
   \label{fig:notjustsize}
\end{figure}

For our experiments, we randomly split each set of images with labels into 80\% for fine-tuning and 20\% for validation. For these 71 training sets, we had a total of approximately 18,000 images: an average of 204 training images and 50 held-out validation images each. There were 5.2 classes per classifier on average, with 2 to 60 classes per classifier.

As seen in Figure~\ref{fig:notjustsize},  the large dataset LC did not always win.  For training datasets which were closer in divergence to ImageNet1K,  the model trained on it was a better base for transfer learning overall. As the task data diverged more and more from ImageNet1K,  LC won more and more.  In Figure~\ref{fig:notjustsize},  the x-axis denotes the divergence of the task data from ImageNet1K  and the y-axis denotes the number of times either LC or ImageNet1K was winner over the 8 learning rate regimes which were tried for each task.  Thus the y-axis values range from 8, denoting when the combined dataset won for all the learning rates tried, to -8, denoting when ImageNet1K won.
%all the learning rates tried.

%Additionally, Figure~\ref{fig:combined2} shows the average accuracy of LC and ImageNet1K for these 71 datasets over the various learning rates regimes tried. As seen, LC does not do better than ImageNet1K over any specific learning rate.

The likely reason is that, although merging datasets certainly increases size, the merged data is also more diverse and tends to have higher divergence.
This empirical result further supports our observations that when considering an individual source dataset, or a merged dataset, or an augmented  source dataset, one needs to carefully consider both indicators of the final performance: the size of the effective source dataset, and its divergence with the target dataset.

\section{Future Work}
The current P2L approach estimates transfer performance at the level of large conceptual categories (e.g.,"animal", or "location"). However, large labeled data sets, such as those used in ImageNet1K, contain deep hierarchies (e.g., animal $\rightarrow$ mammal $\rightarrow$ cat $\rightarrow$ cheetah) that may help to characterize finer resolution maps of the feature space. Identifying crucial sub-features can assist further in selecting more specific source categories, and in developing more efficient source models and transfer techniques.

We currently use only one modality in isolation for determining which source model to use.  However, there is significant other information like accompanying test or audio besides the visual (or the semantic relations) which could additionally aid in determining a good source model. For example, blight is a crop disease and crops are more likely to occur in a plant dataset than any other dataset. If one can determine such links from external datasets, they could help
%narrow in on a good dataset  and 
choose the best source.
%especially when there are two or more close candidates.
Additionally,
extracted tags,
%from the images or using 
or other kinds of semantic information
%available information and using them to find out semantically  closest  source  categories 
extracted from a knowledge graph, can be expected to yield substantial improvements.
%in image recognition.  

Our current P2L method is focused on a single vector characterization of the relationship
%of a single vector representation
of source and target datasets (i.e., similarity).
We plan to extend our study to explore a more complex model, both of the representation of the datasets, as well as oc more enriched relationships. For example, dispersion statistics of the datasets
%divergence or their overlap 
may be provide
%a promising 
further insights into predicting efficacy of the transfer. 

%Additionally, we have currently demonstrated our methods with images and knowledge.  We propose to enhance it towards temporal domains, like machine translation and video.  

\section{Conclusions}

We described an efficient method for fine-tuning a candidate from a family of  pretrained models, applicable to both the image and semantic relations. 
We conducted an empirical test of the method using models trained on specific conceptual categories across images and semantic relations, demonstrating improved transfer learning results, outperforming 
baselines such as  picking the model trained with the largest data set, or a distance measure between source and target or using a common industry standard model like ImageNet1K. These findings suggest that a learned representation from previous tasks can be used to select the best transfer candidate in order to get greater transfer learning. 

Despite order of magnitude differences in training set sizes, we were able to obtain transfer gains by computing an estimate of conceptual closeness. Although prior work has described a saturating curve for training set size contributions to accuracy \cite{Kavzoglu2012}--a curve which we also observed in our data--we showed that feature similarity provided transfer benefits not predicted by dataset size alone. 
%The possibly synergistic contributions of big data and feature similarities remains to be fully explored. 

Our method is efficient at training and classification time, and has been shown to improve accuracy versus the baseline, both on publicly available image and semantic relations datasets. These results help to explain the tension in the literature between results showing that larger datasets usually outperform smaller, \cite{DBLP:journals/corr/RazavianASC14}, but that ill-selected transfer models can nonetheless degrade performance \cite{rosenstein2005transfer}. 

Our results suggest that rather than there being a single ``best'' transfer model, transfer performance critically depends upon the similarity between the source and target models besides the size. Further, methods such as P2L can map the degree of overlap between disparate tasks to select more optimal models.
%and and enhance transfer learning performance. 
Exploring these ``maps'' of feature space similarities could be a valuable future direction for deep learning research.   

%mh: would rather inlcude the Oxford flowers result than this future work
%Our approach estimates transfer performance at the level of large conceptual categories (e.g., ``animal'', or "location"). However, large labeled data sets, such as those used in ImageNet1K, contain deep hierarchies (e.g., animal $\rightarrow$ mammal $\rightarrow$ cat $\rightarrow$ cheetah) that may help to characterize finer resolution maps of the feature space. Identifying crucial sub-features can assist further in selecting more specific source categories, and in developing more efficient source models and transfer techniques.
\newpage

\FloatBarrier
%References and End of Paper
%These lines must be placed at the end of your paper
\bibliography{references}
\bibliographystyle{ieee_fullname}
\end{document}